\pdfoutput=1

\documentclass[11pt]{article}

\usepackage[final]{acl}

\usepackage{times}
\usepackage{latexsym}

\usepackage[T1]{fontenc}

\usepackage[utf8]{inputenc}

\usepackage{microtype}

\usepackage{inconsolata}

\usepackage{graphicx}

\usepackage{hyperref}       
\usepackage{url}            
\usepackage{booktabs}       
\usepackage{amsfonts}       
\usepackage{nicefrac}       
\usepackage{microtype}      
\usepackage{xcolor}         

\usepackage{authblk}
\usepackage{multirow}
\usepackage{amssymb}
\usepackage{dsfont}
\usepackage{amsmath}
\usepackage{enumitem}
\usepackage{natbib}
\usepackage{tabularx}
\usepackage{tcolorbox}
\usepackage{caption}        

\usepackage{enumitem}
\setlist[itemize,1]{left=0pt}
\usepackage{endnotes}

\title{STaR-SQL: Self-Taught Reasoner for Text-to-SQL}

\author{
 \textbf{Mingqian He}\textsuperscript{1},
 \textbf{Yongliang Shen}\textsuperscript{1\textdagger},
 \textbf{Wenqi Zhang}\textsuperscript{1}\\
 \textbf{Qiuying Peng}\textsuperscript{2},
 \textbf{Jun Wang}\textsuperscript{2},
 \textbf{Weiming Lu}\textsuperscript{1}\textsuperscript{\textdagger}
\\
 \textsuperscript{1}Zhejiang University, \textsuperscript{2}OPPO Research Institute
\\
\texttt{\{mingqianhe, syl, zhangwenqi, luwm\}@zju.edu.cn}\\
\texttt{\{pengqiuying, wangjun7\}@oppo.com}
}

\begin{document}
\maketitle
\renewcommand{\thefootnote}{\textdagger}
\footnotetext{Corresponding author.}

\begin{abstract}
Generating step-by-step “chain-of-thought” rationales has proven effective for improving the performance of large language models on complex reasoning tasks. However, applying such techniques to structured tasks, such as text-to-SQL, remains largely unexplored. In this paper, we introduce Self-Taught Reasoner for text-to-SQL (STaR-SQL), a novel approach that reframes SQL query generation as a reasoning-driven process. Our method prompts the LLM to produce detailed reasoning steps for SQL queries and fine-tunes it on rationales that lead to correct outcomes. Unlike traditional methods, STaR-SQL dedicates additional test-time computation to reasoning, thereby positioning LLMs as spontaneous reasoners rather than mere prompt-based agents. To further scale the inference process, we incorporate an outcome-supervised reward model (ORM) as a verifier, which enhances SQL query accuracy. Experimental results on the challenging Spider benchmark demonstrate that STaR-SQL significantly improves text-to-SQL performance, achieving an execution accuracy of 86.6\%. This surpasses a few-shot baseline by 31.6\% and a baseline fine-tuned to predict answers directly by 18.0\%. Additionally, STaR-SQL outperforms agent-like prompting methods that leverage more powerful yet closed-source models such as GPT-4. These findings underscore the potential of reasoning-augmented training for structured tasks and open the door to extending self-improving reasoning models to text-to-SQL generation and beyond.
\end{abstract}

\section{Introduction}
Large Language Models (LLMs) have demonstrated remarkable potential in various language tasks \citep{brown2020language, achiam2023gpt}, including text-to-SQL translation \citep{rajkumar2022evaluating, liu2023comprehensive}. Interacting with complex relational databases typically requires both programming expertise and a deep understanding of the underlying data. Text-to-SQL bridges this gap by allowing non-experts to ask questions in natural language, automatically translating them into SQL queries and returning the results \citep{cai2017encoder, xu2017sqlnet, yaghmazadeh2017sqlizer}. 

\begin{figure*}[ht]
    \centering
    \includegraphics[width=\linewidth]{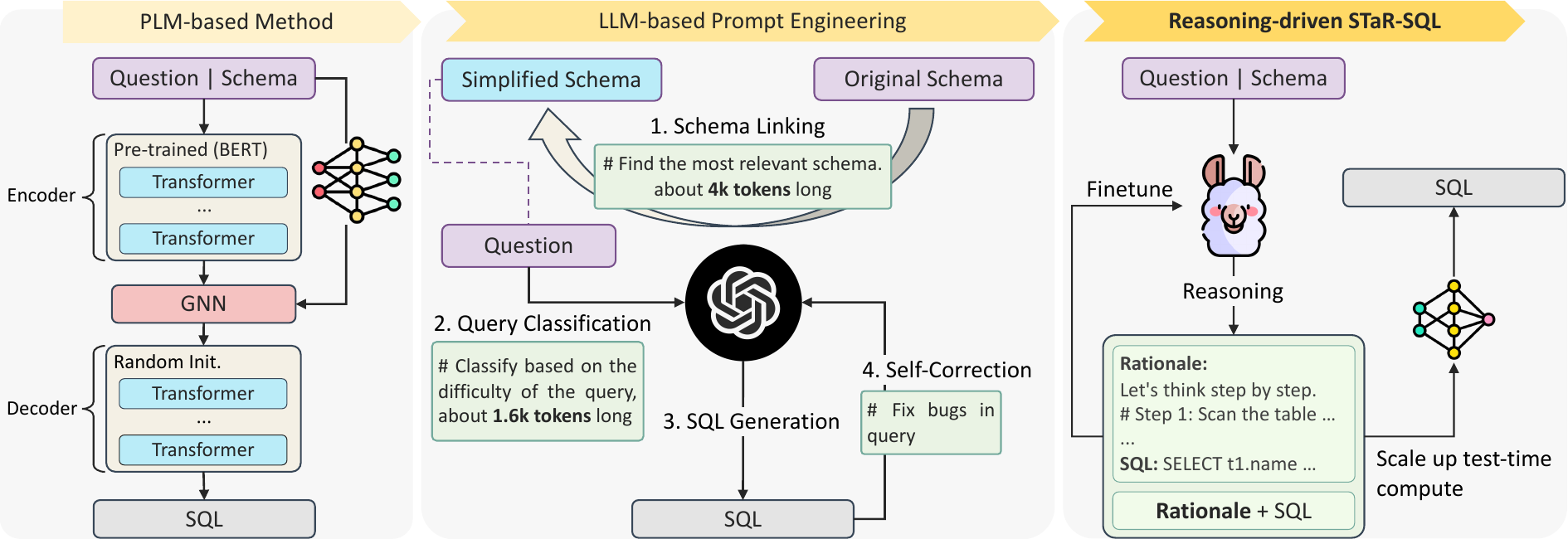} 
    \caption{A comparison of different text-to-SQL methods: Traditional PLM-based methods focus on how to encode the schema (e.g., RATSQL \citep{wang2019rat}). Current LLM-based methods employ carefully designed prompts and subtask flows to simplify and understand the task, functioning in an agent-like manner and using many tokens in the prompt (e.g., DIN-SQL \citep{pourreza2024din}). We treat text-to-SQL as a reasoning-driven process. By leveraging the LLM’s existing reasoning capabilities, we iteratively bootstrap its ability to generate high-quality rationales. In addition, by allocating more test-time computation, we further improve the reliability of the process.}
    \label{fig:intro}
\end{figure*}

Despite significant advancements in this field, most existing approaches primarily harness LLMs for their instruction-following capabilities, focusing on schema selection optimization and result refinement \citep{pourreza2024din}, as illustrated in Figure~\ref{fig:intro}. However, these prompts can be rigid and consume a substantial portion of the available context tokens. Smaller open-source models may also struggle to interpret and follow the carefully crafted prompts on which these methods rely. Moreover, this narrow emphasis on prompt engineering frequently overlooks the powerful reasoning capabilities inherent in LLMs \citep{liu2023evaluating, frieder2024mathematical}. While these methods perform well on simple queries, they tend to falter when confronted with more complex ones \citep{eyal2023semantic}. This shortcoming is particularly problematic for non-experts, who may have trouble verifying whether the generated SQL queries accurately capture their original intent. Complex misalignments in SQL queries can be especially difficult for users to detect and correct.

To address these challenges, we reconceptualize text-to-SQL as a reasoning-driven process, enabling LLMs to handle complex queries by generating step-by-step rationales. This approach offers several key advantages:
\begin{itemize}
    \item \textbf{Robustness for Complex Queries}: A step-by-step chain-of-thought reasoning method enables the model to systematically break down complex queries, handle intricate database schemas more effectively, and produce more accurate results.
    \item \textbf{Scalability through Reasoning}: By allocating additional computational resources at inference time, reasoning performance can be improved. Techniques such as best-of-N sampling \citep{nakano2021webgpt, Askell2021AGL, cobbe2021training} can further boost accuracy.
    \item \textbf{Enhanced Transparency}: Step-by-step rationales provide outputs that are more interpretable and verifiable compared to traditional end-to-end generation approaches.
\end{itemize}

Therefore, we introduce the Self-Taught Reasoner for text-to-SQL (STaR-SQL), a scalable bootstrapping method that enables LLMs to learn to generate high-quality rationales for text-to-SQL. Specifically, we employ few-shot prompting to have a LLM self-generate rationales and then refine its capabilities by fine-tuning on rationales that yield correct answers. To further improve performance on complex queries, we provide the correct answer to the model to guide the generation of useful rationales. These rationales are incorporated into the training data, allowing the model to learn to solve increasingly challenging queries. We repeat this procedure, using the improved model to generate subsequent training sets. Recently, some works have shown that LLMs can leverage additional test-time computation to improve their outputs \citep{snell2024scaling, brown2024large, he2024advancing}. In our experiments, we introduced a verification mechanism to ensure result accuracy by employing an Outcome-supervised Reward Model (ORM) \citep{cobbe2021training, yu2023outcome}, a straightforward yet effective verifier that demonstrably improves overall performance.

We demonstrate the effectiveness of our method on the challenging cross-domain benchmark Spider. Using the two official evaluation metrics (execution accuracy and exact set match accuracy \citep{zhong2020semantic}), our method achieves an execution accuracy of 86.6\%, outperforming both a few-shot baseline (+31.6\%) and a baseline fine-tuned to predict answers directly (+18.0\%). It even surpasses prompting methods \citep{pourreza2024din, gao2023text} that rely on more powerful closed-source models such as GPT-4, setting a new standard for reasoning-driven text-to-SQL approaches.

\section{Related Work}
\subsection{Text-to-SQL}
Text-to-SQL \citep{cai2017encoder, zelle1996learning, xu2017sqlnet, yu2018typesql, yaghmazadeh2017sqlizer}, which aims to convert natural language instructions or questions into SQL queries, has drawn significant attention. Since the work of \citet{dong2016language}, leading text-to-SQL models have adopted attention-based sequence-to-sequence architectures to translate questions and schemas into well-formed SQL queries. These models have increasingly benefited from pre-trained transformer architectures, ranging from BERT \citep{hwang2019comprehensive, lin2020bridging} to larger language models such as T5 \citep{raffel2020exploring} in \citet{scholak2021picard}, OpenAI CodeX \citep{chen2021evaluating}, and GPT variants \citep{rajkumar2022evaluating, liu2023divide, pourreza2024din}. Along with using pre-trained models, various task-specific enhancements have been introduced, including improved schema encoding via more effective representation learning \citep{bogin2019representing} and fine-tuned attention mechanisms for sequence-to-sequence models \citep{wang2019rat}. On the decoding side, some methods incorporate the syntactic structure of SQL \citep{hwang2019comprehensive, xu2017sqlnet, hui2021improving}.

Recent advances in LLMs have also extended their multi-task capabilities to text-to-SQL. In zero-shot scenarios, a task-specific prompt is added before the schema and the question, guiding the LLM to generate an SQL query. \citet{rajkumar2022evaluating, liu2023comprehensive} showed that OpenAI CodeX can achieve 67\% execution accuracy using this approach. Building on this, few-shot prompting strategies have been investigated. In particular, \citet{pourreza2024din, liu2023divide} proposed GPT-4-based DIN-SQL, which divides the problem into four subtasks (schema linking, classification, generation, and self-correction) and achieves strong performance on the Spider benchmark. However, \citet{pourreza2024din} also noted that DIN-SQL encounters difficulties when dealing with complex queries. In contrast to these approaches, our method reframes text-to-SQL as a reasoning task. By doing so, it leverages the inherent reasoning capabilities of LLMs to boost performance and facilitates the integration of additional reasoning techniques into text-to-SQL systems.

\subsection{Multi-step Reasoning}
Complex reasoning tasks have sparked extensive research in LLMs, which are crucial for handling challenging queries \citep{kaddour2023challenges,lightman2023let, huang2023large}. One prominent strategy is the Chain-of-Thought (CoT) prompting technique \citep{wei2022chain}, along with its variants \citep{kojima2022large, wang2022self, yao2024tree}, which decompose the reasoning process into sequential steps and systematically approach problem-solving in a human-like manner. To further enhance the accuracy of these intermediate steps, recent studies leverage extensive synthetic datasets, which are either distilled from cutting-edge models \citep{yu2023metamath, luo2023wizardmath} or composed of self-generated rationales \citep{zelikman2022star, yuan2023scaling, ni2022learning}, to fine-tune the LLMs. Such training strategy effectively sharpens the models' ability to produce correct chain-of-thought reasoning.

Additionally, there is growing interest in test-time verification, which involves generating multiple candidate solutions and ranking them with a separate verifier \citep{cobbe2021training, he2024advancing} to select the most accurate one. For example, the DIVERSE framework \citep{Li2022MakingLM} employs a variety of CoT prompts together with a verifier to address reasoning challenges, while CoRe \citep{zhu2022solving} fine-tunes both the generator and verifier in a dual-process system, improving LLM performance on math word problems.

\begin{figure*}[ht]
 \centering
 \includegraphics[width=\linewidth]{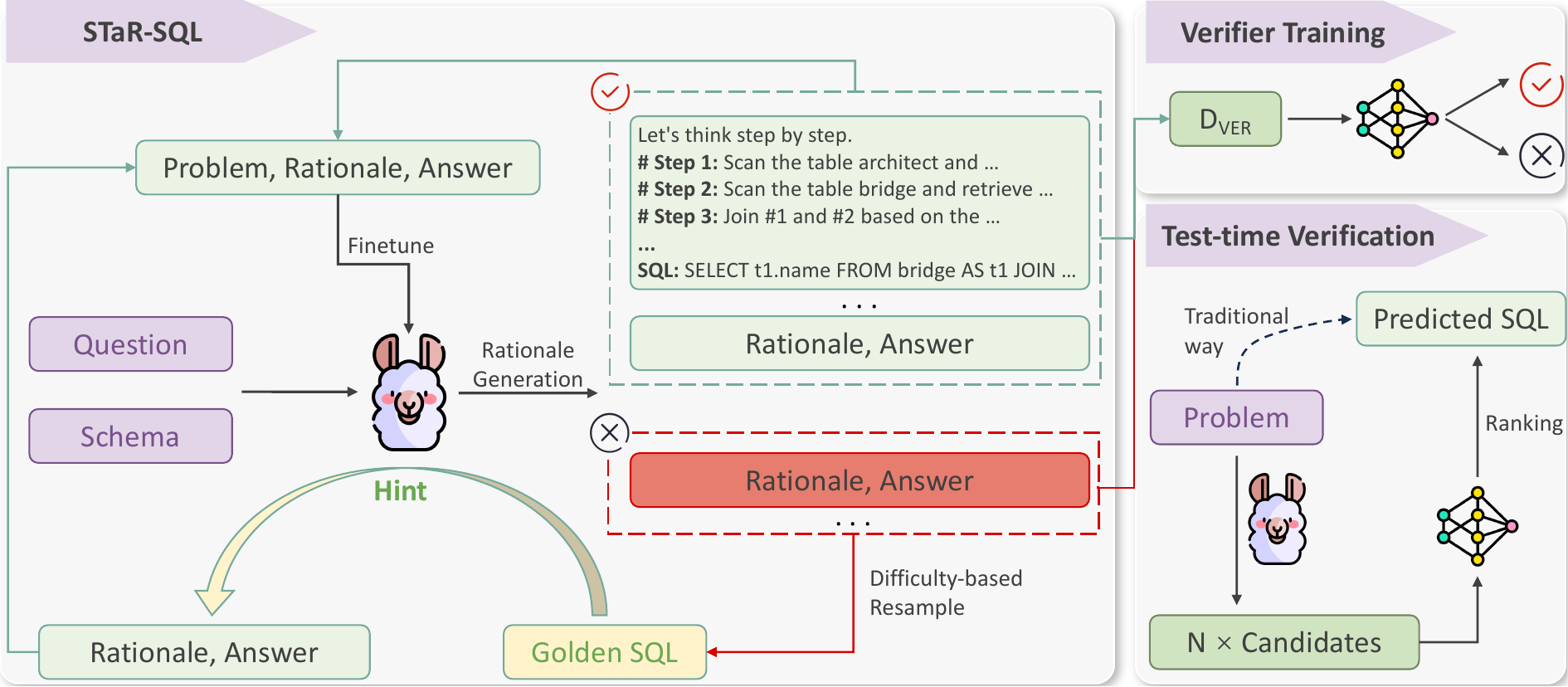}
 \caption{An overview of the STaR-SQL framework. It consists of three main steps: step-by-step rationale generation for self-improvement, verifier training, and test-time verification. We transform text-to-SQL into a reasoning task and further explore scaling up test-time computation by incorporating a verifier and employing best-of-N sampling.}
\end{figure*}

\section{STaR-SQL}
In this section, we introduce STaR-SQL, a method that evokes the intrinsic reasoning capabilities of LLMs to enhance performance on complex text-to-SQL tasks. We begin by describing the problem formulation (§~\ref{3.1}), followed by an explanation of how we generate step-by-step rationales (§~\ref{3.2}) for self-improvement. Finally, we outline our approach to verifier training and scaling up test-time compute to further enhance accuracy (§~\ref{3.3}). A schematic overview of the algorithm is provided in Figure~1.

\subsection{Problem Formulations}
\label{3.1}
The text-to-SQL task involves mapping a question $Q = (q_1, \dots, q_m)$ and a database schema $S = \bigl[ table_1(col_1^1 \dots col_{c_1}^1), \dots, table_T(col_1^T \dots col_{c_T}^T) \bigr]$ to a valid SQL query $Y = (y_1, \dots, y_n)$. Performance is typically evaluated using two metrics: 
1) \textit{exact match}, which compares the predicted query to the golden query in terms of overall structure and within each field token by token, and 
2) \textit{execution match}, which checks whether the prediction produces the same results as the golden query when executed on the database.

\subsection{Self-Taught Reasoner}
\label{3.2}
Self-Taught Reasoner (STaR; \citet{zelikman2022star}) is an iterative approach in which a language model improves itself using correctness feedback. We begin with a pre-trained LLM $\pi_\theta$ as a generator and an initial text-to-SQL dataset $\mathcal{D}=\{(Q_i, S_i, Y_i)\}_{i=1}^D,$ where each instance comprises a question $Q_i$, a database schema $S_i$, and a corresponding golden SQL query $Y_i$. Our method also assumes a small prompt set $\mathcal{P}$ of examples with intermediate rationales $R$: $\mathcal{P} = \{(Q_i^p, S_i^p, R_i^p, Y_i^p)\}_{i=1}^P$, where $P \ll D$ (for instance, $P=3$). Following the standard few-shot prompting procedure, we concatenate this prompt set $\mathcal{P}$ to each example in $\mathcal{D}$, then sample $k$ rationales followed by an answer from the generator: $\{(R_i^j, \hat{Y}_i^j) \sim \pi_\theta(R, \hat{Y} | \mathcal{P}, Q_i, S_i)\}_{j=1}^{k}$.

Having access to golden SQL queries $Y_i$, we can assign a binary correctness label to each generated query $\hat{Y}_i^j$ using the indicator $\mathds{1}[\hat{Y} = Y]$. A rationale is labeled as correct if its final query $\hat{Y}$ matches the golden query $Y$. Intuitively, correct queries should stem from higher-quality rationales, so we only retain those correct rationales. However, under these conditions, models tend to over-sample solutions for simpler queries while under-sampling solutions for more complex queries, a phenomenon known as tail narrowing \citep{ding2024mitigating}. This results in a training set for the next iteration dominated by rationales for simpler problems, with limited coverage of more challenging queries, thereby introducing sampling bias. 

To address this issue, we employ a straightforward difficulty-based resampling strategy, which has proven sufficiently effective in practice. Specifically, for each question, we resample $L$ times, where $L$ is the number of incorrect initial responses for that question. To improve accuracy, we provide the golden SQL query as a hint to the model and ask it to generate rationales in the same style as during the previous rationale-generation step. Given the golden SQL query, the model can more easily reason backwards to produce a rationale that yields the correct answer. For correct initial responses, we directly add them to the training set.

We then form a new dataset, $\mathcal{D}_{\text{SFT}}$, and perform supervised fine-tuning (SFT) of the generator \(\pi_\theta\) using the negative log-likelihood objective:
\begin{equation}
    \mathcal{L}_{\rm{SFT}} = -\mathds{E}_{(X, R, Y) \sim \mathcal{D}_{\rm{SFT}}} \sum_{i=1}^{|R|+|Y|} \log \pi_\theta(t_i|t_{<i}, X)
\end{equation}
where $X$ is the concatenation of the question $Q$ and the schema $S$, i.e., $X=(Q, S)$.

The newly fine-tuned generator is used in subsequent iterations. Once we collect a new dataset, we always return to the \emph{original} pre-trained model $\pi_\theta$ for re-initialization (as opposed to continually fine-tuning the same model) to mitigate overfitting. This process is repeated until performance plateaus.

\subsection{Test-time verification}
\label{3.3}
Previous self-improvement methods such as RFT \citep{yuan2023scaling}, STaR, and ReST \citep{gulcehre2023reinforced} typically discard incorrect model-generated solutions. However, even incorrect solutions can contain useful information: a language model may learn from the discrepancies between correct and incorrect solutions, identifying common error patterns and thereby improving its overall accuracy. In this work, we propose utilizing both correct and incorrect solutions in the iterative process to train a verifier. Following \citet{cobbe2021training}, we introduce a verifier, also known as an outcome-supervised reward model (ORM). An ORM estimates the probability that a candidate rationale $T$ is correct for a given problem. It is built upon a LLM with an additional randomly initialized linear layer that outputs a scalar value. The ORM is trained with a binary classification loss:
\begin{equation}
    \mathcal{L}_{\rm{ORM}} = A_T \log r_T + (1-A_T) \log (1-r_T)
\end{equation}
where $A_T$ is the correctness label ($A_T = 1$ if $T$ is correct, otherwise $A_T = 0$), and $r_T$ is the ORM's sigmoid output. In our context, $A_T$ is defined by the execution match label; i.e., whether the generated SQL query matches the golden query when executed. Since each generated rationale is labeled during every iteration, these labeled pairs form an ideal training set $\mathcal{D}_{\text{VER}}$ for the verifier.

We further scale up test-time compute through \emph{best-of-$N$} sampling strategy \citep{nakano2021webgpt, Askell2021AGL, cobbe2021training}, which improves the reliability of the final answer. Specifically, at test time, the language model generates \(N\) candidate solutions in parallel, and the one with the highest verifier score is chosen as the final output.

\section{Experiments}
\subsection{Experimental Setup}
\paragraph{Datasets}
Several large text-to-SQL datasets have been created, some with single schemas \citep{wang2019rat} or with simple queries \citep{zhong2017seq2sql}. Notably, the Spider dataset \citep{yu2018spider} consists of 10,181 questions and 5,693 unique complex SQL queries across 200 databases, covering 138 domains, each containing multiple tables. The standard protocol for this dataset divides it into 8,659 training examples across 146 databases and 1,034 development examples across 20 databases, with non-overlapping databases in each set. SQL queries are categorized into four difficulty levels, based on the number of SQL keywords used, the presence of nested subqueries, and the usage of column selections and aggregations. The dataset is used to assess the generalization capabilities of text-to-SQL models on complex queries with unseen schemas. We focus on this dataset for our experiments, as it enables comparison with many previous methods.

\paragraph{Metrics}
The performance of our models are evaluated using the official metrics of Spider \citep{zhong2020semantic}: exact-setmatch accuracy (EM) and execution accuracy (EX). The exact-set-match accuracy (EM) treats each clause as a set and compares the prediction for each clause to its corresponding clause in the reference query. A predicted SQL query is considered correct only if all of its components match the ground truth. This metric does not take values into account. The execution accuracy (EX) compares the execution output of the predicted SQL query with that of the ground truth SQL query on some database instances. Execution accuracy provides a more precise estimate of the model’s performance since there may be multiple valid SQL queries for a given question, and exact set match accuracy only evaluates the predicted SQL against one of them.

\paragraph{Parameter Setting}
We used Llama-3.1-8B-Instruct as our base language model. This open-source model demonstrates non-trivial performance on the text-to-SQL task while leaving room for further improvements, making it an ideal testbed for our study. To construct the training dataset, we selected 7,000 problems from the Spider training set and sampled 8 solutions for each problem. We then filtered the correct solutions to train the generator and used the entire dataset to train the verifier. We ran STaR-SQL until performance plateaued and report the best results observed.

\paragraph{Baselines}
We conducted a comparative evaluation against several well-established methods, including traditional pre-trained transformer-based models (PLM-based) that directly predict SQL or intermediary representations. For LLM-based methods, we compared STaR-SQL with several notable prompt-engineering approaches utilizing strong closed-source LLMs, with particular emphasis on DAIL-SQL \cite{gao2023text}, which is currently the SOTA approach of this kind. We also compared our method with fine-tuned specialized code LLMs, such as CodeS \citep{li2024codes}, DTS \citep{pourreza2024dts} and ROUTE \citep{qin2024route}. Regarding training data generation, we considered Question Decomposition (QD) \cite{eyal2023semantic} as a baseline. In this approach, the model is instructed to first produce a custom intermediary language, QPL, which is then translated into the rationale. To assess data quality, we compared a model trained on QD-generated data with our own approach. Finally, we included an LLM fine-tuned to predict answers directly, without revealing its reasoning steps, to demonstrate the importance of incorporating a reasoning process.

\begin{table*}[ht]
 \small
 \centering
 \begin{tabular}{l l l c c}
  \toprule
  \textbf{Classification} & \textbf{Methods} & \textbf{Models} & \textbf{EX} & \textbf{EM} \\
  \midrule
  \multirow[t]{3}{*}{PLM-based} & NatSQL \cite{gan2021natural} & RAT-SQL \cite{wang2019rat} & 73.7 & - \\
                                & QPL \cite{eyal2023semantic} & Flan-T5-XL \cite{chung2024scaling} & 77.4 & - \\
                                & Graphix-T5 \cite{li2023graphix} & Graphix-T5 & 78.2 & \bf{75.6} \\
  \midrule
  \multirow[t]{5}{*}{Prompting with LLMs} & \multirow[t]{5}{*}{Few-shot} & Llama-3.1-8B-Instruct & 55.0 & 34.2 \\
                                &  & Qwen2.5-7B \citep{yang2024qwen2} & 72.5 & - \\
                                &  & CodeX Cushman & 43.1 & 30.9 \\
                                &  & CodeX Davinci & 61.5 & 50.2 \\
                                &  & GPT-4 & 67.4 & 54.3 \\
                                \cmidrule(lr){2-5}
                                & \multirow[t]{2}{*}{DIN-SQL} & Llama-3.1-8B-Instruct & 45.2 & 26.5 \\
                                & \cite{pourreza2024din} & GPT-4 & 74.2 & 60.1 \\
                                \cmidrule(lr){2-5}
                                & \multirow[t]{2}{*}{MAC-SQL} & Llama-3-8B & 64.3 & - \\
                                & \cite{wang2023mac} & Qwen2.5-7B & 71.7 & - \\
                                \cmidrule(lr){2-5}
                                & \multirow[t]{2}{*}{MCP \citep{qin2024route}} & Llama-3-8B & 75.0 & - \\
                                &  & Qwen2.5-7B & 78.3 & - \\
                                \cmidrule(lr){2-5}
                                & \multirow[t]{2}{*}{DAIL-SQL} \cite{gao2023text} & GPT-3.5-TURBO & 77.8 & 63.9 \\
                                &  & GPT-4 & 81.7 & 69.1 \\
    \midrule
    \multirow[t]{1}{*}{Fine-Tuning} & predict SQL-only & Llama-3.1-8B-Instruct & 68.6 & 57.9 \\
    \multirow[t]{4}{*}{with Open-Source LLMs} & QD \cite{eyal2023semantic} & Llama-3.1-8B-Instruct & 64.5 & 54.3 \\
                                & CodeS \citep{li2024codes} & StarCoder & 69.8 & - \\
                                & DTS-SQL \citep{pourreza2024dts} & Mistral-7B & 77.1 & 69.3 \\
                                & SENSE-7B \citep{yang2024synthesizing} & CodeLlama-7B & 83.2 & - \\
                                & ROUTE \citep{qin2024route} & Qwen2.5-7B & \underline{83.6} & - \\
                                & STaR-SQL & Llama-3.1-8B-Instruct & 75.0 & 64.9 \\
                                & STaR-SQL ORM@16 & Llama-3.1-8B-Instruct & \bf{86.6} & \underline{72.5} \\
  \bottomrule
 \end{tabular}
 \caption{Execution accuracy (EX) and exact set match accuracy (EM) (both in \%) on the dev set of Spider. \textbf{Bold} indicates the best results, and \underline{underline} indicates the second best.}
 \label{tab:mainrst}
\end{table*}

\subsection{Main Results}
Most of our evaluation during development was conducted on the Spider development set, which was easily accessible, unlike the test set that was only accessible through the evaluation server provided by \citet{yu2018spider}. As shown in Tables~\ref{tab:mainrst}, our proposed method significantly enhances the original performance of Llama-3.1-8B-Instruct, improving its accuracy from 55.0\% to 75.0\% (+20.0\%). Although small open-source models cannot directly apply reasoning to the text-to-SQL task and thus perform poorly, they demonstrate the potential to employ reasoning abilities when trained on correct rationales. Our approach also outperforms naive few-shot prompting methods, showing that it is crucial for LLMs to be familiar with the reasoning patterns required for this task: STaR-SQL surpasses few-shot prompting with stronger closed-source LLMs like GPT-4 by a large margin (+7.6\%), and it is comparable to advanced prompt engineering techniques and specialized code LLMs like CodeS and DTS-SQL. Notably, it even outperforms DIN-SQL, which relies on extensive compute to simplify schemas and refine the output. Compared to predicting only the final SQL, our results demonstrate the necessity of integrating the reasoning process during inference, as this improves accuracy by an additional 6.4\%.

When we scale up test-time compute, the benefit of reframing the text-to-SQL task as a reasoning process becomes even more evident. By sampling 16 solutions for each problem and applying ORM for selection, our approach significantly surpasses other PLM-based and LLM-based methods in terms of exact set match. For example, it achieves the highest accuracy of 86.6\%, outperforming DAIL-SQL (the best GPT-4 prompting method) by 4.9\% and the previous state-of-the-art ROUTE by 3.0\%. Furthermore, training ORM does not require additional data because it is derived entirely from STaR-SQL's iterative training process. As a result, this method is both data-efficient and straightforward, leveraging both correct and incorrect solutions from an iteratively trained generator to build a robust verifier. These results highlight STaR-SQL’s strong performance and scalability when increasing test-time compute.

\begin{figure*}[!ht]
    \centering
    \includegraphics[width=0.98\linewidth]{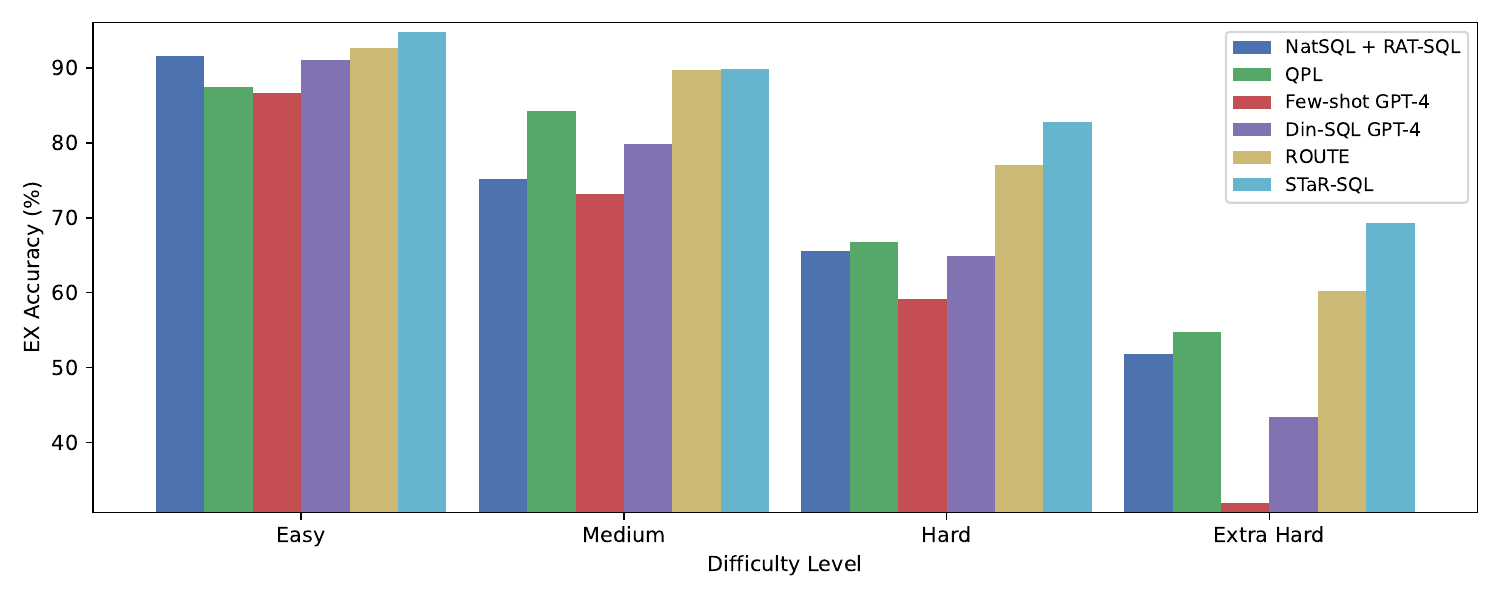}
    \caption{Execution accuracy comparison across different query difficulty levels on the Spider development set.}
    \label{fig:1_difficult}
\end{figure*}

We attribute these improvements to the following factors: 1) Reasoning Integration: Beyond leveraging the large language model's understanding capability, we also utilize its reasoning ability during inference. This transforms the model from a mere “agent” into a “reasoner,” enabling it to handle complex query problems more effectively. 2) Expanded Test-Time Computation: We scale up test-time computation, which complements our approach of reframing text-to-SQL as a reasoning task. Allocating more computational resources proves to be an effective way to boost performance. 3) Learning from Errors: Our method also learns from the model’s own erroneous reasoning rationales by using ORM as guidance. This strategy improves the accuracy of generation while maintaining data efficiency.

\subsection{Execution Accuracy by Difficulty Level}
We further analyzed the performance of our method on queries of varying difficulty. Figure~\ref{fig:1_difficult} compares our approach with basic few-shot prompting and other advanced techniques on the Spider development set, demonstrating that our method consistently outperforms all baselines across every difficulty level. Although these competing methods often exceed 90\% accuracy on easy queries, their performance can drop to approximately 50\% on more complex ones—even specialized code LLMs fare poorly in such scenarios. This decline is particularly problematic for non-experts, who may struggle to verify whether a complex SQL query matches their intended question \citep{eyal2023semantic}. Notably, our method achieves the greatest gains in the extra-hard (69.3\%) and hard (82.8\%) categories, outperforming the second-best results by +5.8\% and +9.1\%, respectively. These gains stem from integrating reasoning into the inference process, leveraging the model’s reasoning capabilities to address complex queries, and highlighting the importance of shifting the problem-solving paradigm.
\begin{figure}[!ht]
 \centering
 \includegraphics[width=\linewidth]{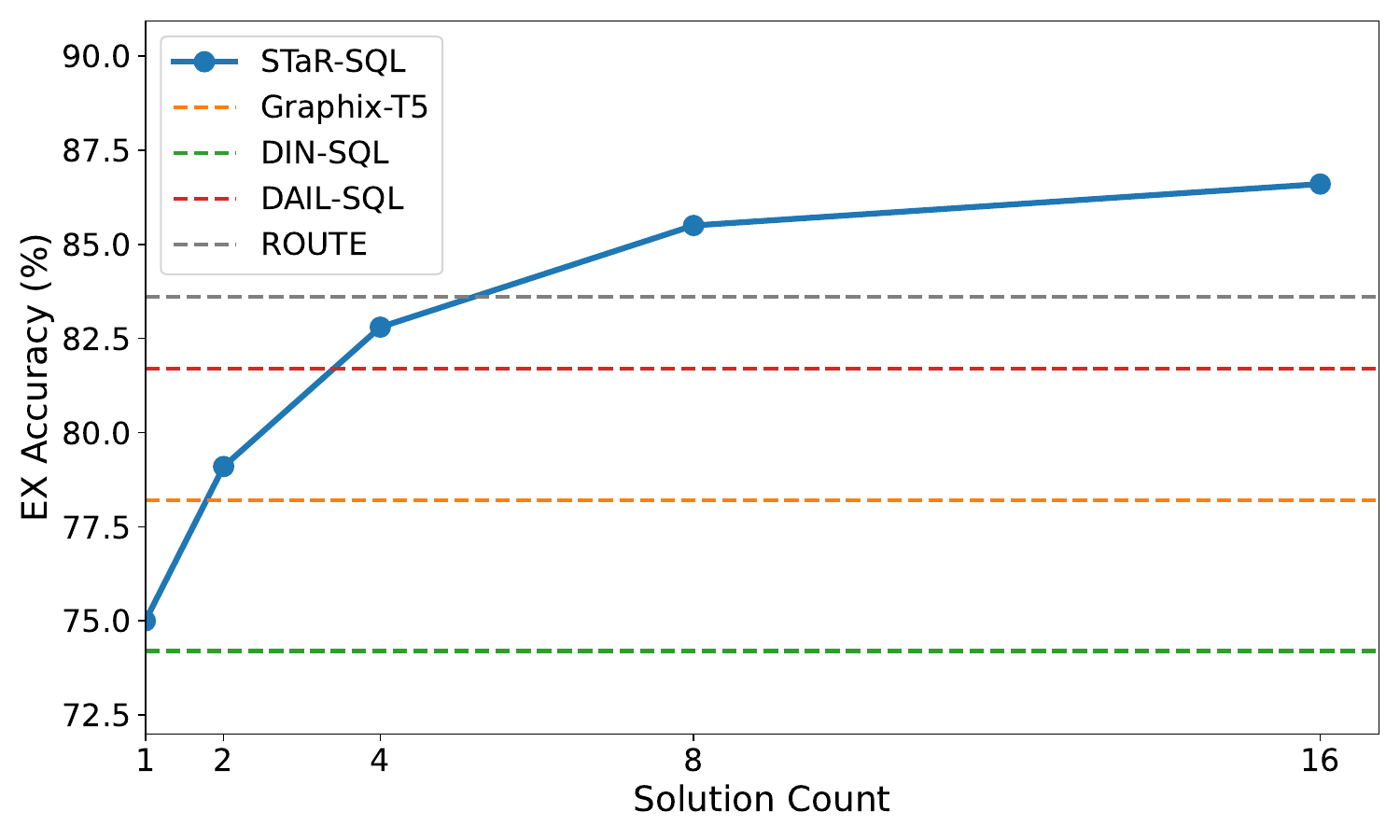}
 \caption{Performance of STaR-SQL with varying numbers of solutions (N).}
 \label{fig:2_num_cand}
\end{figure}

\subsection{Different Amounts of Candidate Solutions}
The number of candidate solutions affects verification performance. While a larger pool of solutions can introduce additional, potentially superior candidates, it also increases computational overhead and may lead to diminishing returns. In our study, we restrict the maximum number of solutions to 16. As shown in Figure~\ref{fig:2_num_cand}, increasing the number of samples consistently improves performance. Notably, sampling 4 solutions already enables STaR-SQL to surpass the best prompt-engineering method, DAIL-SQL, which depends on the more powerful but closed-source GPT-4. With 8 solutions, STaR-SQL further outperforms the state-of-the-art specialized code LLM, ROUTE, by 1.9\%. These results demonstrate that substantial accuracy gains can be achieved with only a slight increase in test-time computation. Our findings align with recent work suggesting that increased test-time compute enhances reasoning performance \citep{snell2024scaling, brown2024large, wu2024inference}. Moreover, allocating additional tokens to the reasoning process, rather than to carefully engineered prompts, proves more effective. For example, our method achieves a 41.4\% improvement over DIN-SQL, which uses more than 6k tokens in its prompt.
\begin{figure}[!ht]
 \centering
 \includegraphics[width=\linewidth]{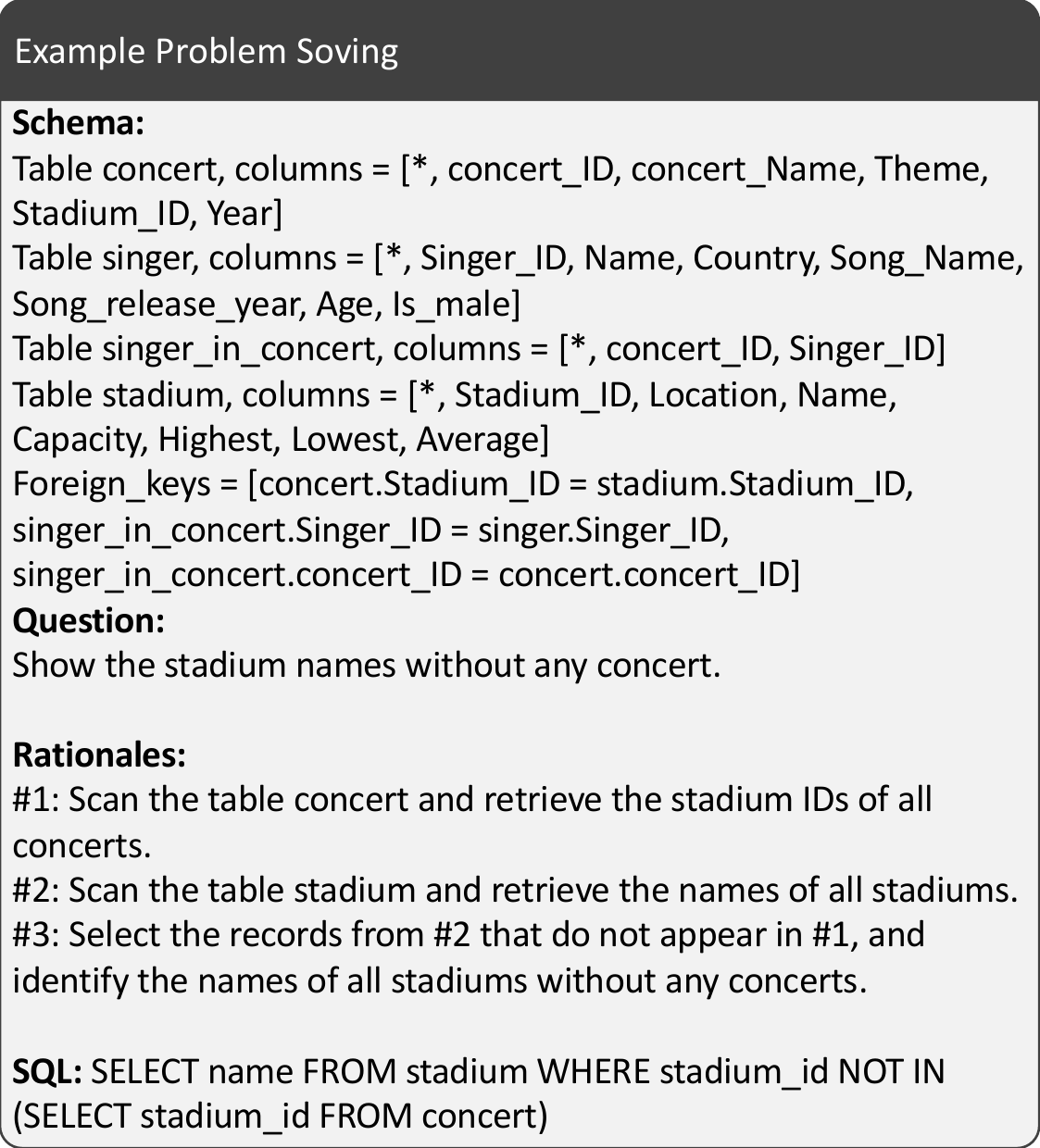}
 \caption{A case study from the Spider dev set.}
 \label{fig:case_study}
\end{figure}

\subsection{Case Study}
We also conduct a case study to intuitively demonstrate the effectiveness of STaR-SQL. As shown in Figure~\ref{fig:case_study}, when confronted with a complex question, STaR-SQL successfully decomposes the problem into a series of reasoning steps, progressively guiding the generation of the final SQL query. In addition, STaR-SQL enhances transparency by presenting the entire query generation process and providing a clear rationale for the final result. This transparency not only improves interpretability but also enables users to verify whether the generated query aligns with their intended question, making it easier to validate consistency between the input and output compared to other methods.

\subsection{Ablation Study}
We conduct an ablation study to evaluate three key components of our framework: (a) the use of intermediate rationales (step-by-step reasoning), (b) the best-of-N sampling strategy during inference, and (c) the verifier-based ranking compared to a self-consistency (majority voting) baseline. Table~\ref{tab:ablation} summarizes the results under different settings. We observe that: 1) Removing step-by-step reasoning severely degrades both execution accuracy (EX) and exact match (EM), underscoring the necessity of intermediate reasoning. 2) Omitting best-of-N sampling reduces accuracy, highlighting the benefit of scaling test-time computation. 3) Replacing the verifier with self-consistency improves performance over single-shot generation but still falls short of our verifier-based approach.

\begin{table}[!ht]
    \centering
    \small
    \begin{tabular}{lcc}
        \toprule
        \textbf{Method} & \textbf{EX} & \textbf{EM} \\
        \midrule
        Ours             & \bf{86.6} & \bf{72.5} \\
        w/o rationales   & 68.6 & 57.9 \\
        w/o best-of-N    & 75.0 & 64.9 \\
        Self-Consistency & 78.8 & 71.7 \\
        \bottomrule
    \end{tabular}
    \caption{Results of the ablation study, demonstrating the impact of different components of STaR-SQL.}
    \label{tab:ablation}
\end{table}

\section{Conclusion}
In this paper, we propose STaR-SQL, an innovative method that leverages the intrinsic reasoning capabilities of language models to perform step-by-step reasoning for text-to-SQL problems. We iteratively bootstrap the ability to generate high-quality rationales and integrate a verifier to enhance the accuracy. Our empirical findings highlight the efficacy of STaR-SQL: our model achieves state-of-the-art results among fine-tuned models on the Spider dev set (without database values), especially on hard and extra-hard queries, demonstrating notable performance improvements over existing PLM-based and LLM-based methods. Through step-by-step reasoning, the large language model makes the entire process more interpretable than merely generating SQL or intermediate representations—particularly for complex queries. At the same time, by allocating additional test-time computation, we further improve accuracy, illustrating the scalability and potential of our method.

In future work, we plan to explore more effective ways of utilizing test-time compute to boost the reasoning capabilities of language models on text-to-SQL tasks. We have begun experimenting with a stronger verifier—a process-supervised reward model (PRM)—which employs fine-grained supervision signals. Beyond the best-of-N approach, there are also other methods for using test-time compute to enhance LLM performance. For instance, one can modify the proposal distribution for responses by prompting the model to sequentially revise its outputs, or alter how the verifier is used (e.g., leveraging Monte Carlo Tree Search or other search strategies). We believe these directions hold promise for further improving the robustness and accuracy of text-to-SQL systems.

\clearpage
\section*{Limitations}
Although STaR-SQL is effective for text-to-SQL tasks under simple schema encoding, it remains uncertain whether additional methods for rich schema encoding could further enhance performance. As our approach transforms text-to-SQL into a reasoning task, we have not yet integrated techniques to improve reasoning, such as using more powerful verifiers like process-supervised reward models (PRMs) or search strategies like Monte Carlo Tree Search (MCTS). Addressing these considerations will be the focus of our future research.

\section*{Ethics Statement}
The development of STaR-SQL aims to improve the accuracy and reliability of text-to-SQL tasks using Large Language Models (LLMs). While our method poses no immediate ethical concerns, we acknowledge the potential for misuse if applied in sensitive areas such as automated decision-making. We recommend rigorous evaluation and oversight to prevent bias and ensure data privacy in all applications. Transparency and adherence to ethical standards are crucial in the deployment of these technologies.

\bibliography{custom}

\end{document}